\RequirePackage{fix-cm}
\documentclass[smallextended]{svjour3}       
\smartqed  

\usepackage{graphicx}
\usepackage{epstopdf}
\usepackage{subfigure}
\usepackage{amssymb}
\usepackage{mathtools}
\usepackage{multirow}
\usepackage{booktabs,siunitx}
\begin{document}

\title{Fingerprint liveness detection using local quality features 
}


\author{Ram Prakash Sharma \and
        Somnath Dey 
}


\institute{ Ram Prakash Sharma\\
              \email{phd1501201003@iiti.ac.in}\\
         \and
           Somnath Dey\\
           \email{somnathd@iiti.ac.in}\\
           Indian Institute of Technology
           Indore, India}

\date{Received: date / Accepted: date}

\maketitle

\begin{abstract}
Fingerprint-based recognition has been widely deployed in various applications. However, current recognition systems are vulnerable to spoofing attacks which make use of an artificial replica of a fingerprint to deceive the sensors. In such scenarios, fingerprint liveness detection ensures the actual presence of a real legitimate fingerprint in contrast to a fake self-manufactured synthetic sample. In this paper, we propose a static software-based approach using quality features to detect the liveness in a fingerprint. We have extracted features from a single fingerprint image to overcome the issues faced in dynamic software-based approaches which require longer computational time and user cooperation. The proposed system extracts 8 sensor independent quality features on a local level containing minute details of the ridge-valley structure of real and fake fingerprints. These local quality features constitutes a 13-dimensional feature vector. The system is tested on a publically available dataset of LivDet 2009 competition. The experimental results exhibit supremacy of the proposed method over current state-of-the-art approaches providing least average classification error of 5.3\% for LivDet 2009. Additionally, effectiveness of the best performing features over LivDet 2009 is evaluated on the latest LivDet 2015 dataset which contain fingerprints fabricated using unknown spoof materials. An average classification error rate of $4.22 \%$ is achieved in comparison with $4.49 \%$ obtained by the LivDet 2015 winner. Further, the proposed system utilizes a single fingerprint image, which results in faster implications and makes it more user-friendly.
\end{abstract}

\vspace{-3mm}
\section{Introduction}
\vspace{-3mm}
Biometrics-based authentication has drawn the attention of the researchers due to its widespread applications in security and access control. In particular, fingerprint-based authentication is the most widely adopted for personal identification due to its uniqueness and ease in the acquisition. However, fingerprint-based recognition systems are vulnerable to presentation attack using an artificial replica of a fingerprint image. These artificial replica can be made of various materials, i.e., silicone, gelatin, playdoh, etc. Therefore, a suitable countermeasure should be developed to protect fingerprint recognition systems. Liveness detection is an efficient way to circumvent these presentation attacks. In fingerprint recognition system, liveness detection determines whether the fingerprint is genuine or fake. Liveness detection methods can be classified into two categories, i.e., hardware-based or software-based methods. Hardware-based methods utilize additional hardware to measure temperature of the finger, the electrical conductivity of the skin and pulse oximetry. However, the usage of an additional hardware increases the overall cost of the recognition system and requires interaction of the end user with the extra hardware. On the other hand, software-based methods process the biometric sample to detect the vitality information directly from the fingerprint images. Hence, we emphasize on the software-based liveness detection system.

The existing software-based approaches use perspiration based features \cite{Abhyankar2006,Derakhshani2003,MARASCO2012,NIKAM2009,Nikam2010} pore based features \cite{Choi2007,Espinoza2011,Manivanan2010}, quality based features \cite{Choi2009,GALBALLY2012,Galbally2014,Nikam2008} for liveness analysis. Perspiration based features can be lost if the finger pressure is not applied correctly or uniformly while pore based features requires high resolution images for feature extraction. The existing quality based fingerprint liveness detection methods use a single feature or multiple feature based approach. Different sensors capture information differently when used with various materials such as silicone, playdoh, latex, and wood-glue for fake fingerprint fabrication. In such scenarios, a single feature is insufficient to perform well over the different materials used for fabrication. The existing multiple quality features based approaches \cite{GALBALLY2012,Galbally2014} are not able to perform well over different types of sensors. Therefore, the main objective of our work is to propose a novel set of quality related features which can perform equally well for different sensors. The next objective is to evaluate the contribution of the proposed quality based features including few existing quality features to improve the performance of the liveness detection system. Finding the overall best performing features for fingerprint liveness detection is another objectives of the proposed approach.

In this paper, we introduce a set of novel quality based features to alleviate the limitations of existing static software-based approaches where a single or multiple features may fail to detect fake fingerprints fabricated using different materials and
sensors. Further, the proposed approach requires only single image to extract all quality features. The fake fingerprint comprises  different artifacts in the ridge-valley structure owing to elastic characteristics of the materials (gelatin, playdoh, and silicon, etc.) used for fabrication. The proposed method assesses the ridge-valley structure of real and fake fingerprints and extracts features i.e. ridge width smoothness ($RWS$), valley width smoothness ($VWS$), number of abnormal ridges ($R_{ab}$), number of abnormal valleys ($V_{ab}$), ridge valley clarity ($RVC$), frequency domain analysis ($FDA$), orientation certainty level ($OCL$), Gabor quality ($G$) on a local level (block-wise). Thereafter, feature selection unit chooses the best feature set for each fingerprint sensor. The selected feature set is fed to random forest classifier to identify the fingerprint images as real or fake. Experimental results of the proposed method outperform the state-of-the-art for LivDet 2009 datasets. Performance of the best performing features over LivDet 2009 is evaluated on LivDet 2015 datasets which contains fingerprints fabricated using unknown spoof materials and captured with different sensors. The experimental results certify the suitability of the best performing features for different sensors and unknown fake fingerprint fabrication materials.

In a nutshell, the major contributions are summarized as follows.
\begin{itemize}
	\item In this work, a novel set of quality features (RWS, VWS, $R_{ab}$, $V_{ab}$, RVC) are proposed to detect fingerprint liveness on a local level from  single image.
	\item The proposed method investigates joint contribution of proposed quality features including few existing quality features for fingerprint liveness detection.
	\item In order to use local features effectively, feature variance across different blocks of the fingerprint image and overall mean of the local feature values is used to differentiate live or fake fingerprints.
	\item Significant performance improvement has been achieved utilizing the best subset of features for LivDet 2009 datasets. 			
	\item Performance of the best performing features over LivDet 2009 is tested on LivDet 2015. Experimental results affirms suitability of the best performing features for unknown spoof materials.
\end{itemize}

The rest of the paper is organized as follows. In section 2, an overview of the related works for fingerprint liveness detection is presented. Section 3 describes the proposed method for fingerprint liveness detection using quality-related features. The experimental results are thoroughly discussed and compared with the current state-of-the-art in section 4. Finally, the conclusions are drawn in section 5.
\vspace{-5mm}
\section{Related works}
\vspace{-2mm}
In this section, we describe major software-based approaches that have been proposed in the previous studies.

An initial study in the field of fingerprint liveness detection is proposed by Derakhshani et al. \cite{Derakhshani2003}. They have initiated the fingerprint liveness detection research utilizing the skin perspiration phenomenon. In their approach, they have used the periodicity of sweat and sweat diffusion pattern using ridge signals extracted from the fingerprints to identify fake or live fingerprint image. Abhayankar et al. \cite{ABHY2009,ABHYA2004} proposed a wavelet-based method to detect fingerprint liveness. They have determined the liveness of fingerprints using perspiration phenomenon, which changes along the fingerprint ridges in live fingerprints. The changing perspiration pattern is considered as distinctive spatial property, which results from the physical surface properties such
as sweat pore pressure, positioning, roughness of fingerprint,
and so forth. In \cite{Tan2006}, Tan et al. proposed an intensity based perspiration detection approach, which quantifies the grey level differences using histogram distribution statistic to distinguish live or fake fingerprints. Tan et al. \cite{Tan2008} have also proposed fingerprint valley noised based approach as live fingerprints have a clear ridge-valley structure unlike fake fingerprints which have a distinct noise distribution due to the material’s properties. DeCann et al. \cite{DeCann2009} proposed  a novel perspiration detection method which quantifies perspiration via region labeling. Marasco et al. \cite{MARASCO2012} proposed a method combining texture features and skin perspiration patterns. In their study, experiments were carried on standard LivDet 2009 which produces an overall accuracy rate of 74.4 \% over three senors (Biometrika, Crossmatch, and Identix).
Nikam et al. proposed a curvelet \cite{NikamConf2008,NIKAM2009} and ridgelet \cite{Nikam2010} transform based method to detect fingerprint liveness which represents singularities along ridge lines in a more efficient way than the wavelets. Energy and co-occurrence signatures od ridgelet and curvelet are used to distinguish live and fake fingerprint textures. The classifier namely, neural network, support vector machine, and k-nearest neighbor along with one ensemble classifier was used to detect real and fake fingerprints of their custom-made database. In perspiration based approaches, features discriminating the live or fake fingers can be lost if the pressure is not applied correctly or finger is not kept for a fixed amount of time on the sensor. In addition, it requires more user cooperation to capture multiple fingerprint images and cannot be used for real time applications. 

Manivanan et al. \cite{Manivanan2010} proposed the use of sweat pores to detect fingerprint liveness. In their work \cite{Manivanan2010}, they have utilized highpass filtering followed by correlation filter to extract and locate active sweat pores in a fingerprint image. Espinoza et al. \cite{Espinoza2011} proposed an approach utilizing a number of sweat pores. They have used difference in the number of sweat pores as a basis for fingerprint liveness detection. Choi  et al. \cite{Choi2007} makes use of sweat pores spacing and distance for fingerprint liveness detection. According to the authors, the periodicity of a pore in a live fingerprint can be detected more accurately when a finger is dry. In \cite{Marcialis2010}, Marcialis et al. proposed use of pores distribution in order to discriminate between fake and live fingerprint images. They claimed that frequency of pores in live fingerprint is less than that in fake fingerprints, due to fabrication steps necessary for replica. The major limitation of pore based approaches is that, it requires high resolution images to detect sweat pores accurately.

Moon et al. \cite{Moon2005} proposed a wavelet-based method analyzing texture coarseness difference between real and fake fingerprints for liveness detection. Abhyankar et al. \cite{Abhyankar2006} proposed a method using multi-resolution texture analysis along with inter-ridge frequency analysis to distinguish live fingers. In this work \cite{Abhyankar2006}, the detection depends on the characteristics of underlying fingerprint texture, which are different for live and fake fingerprints. Lee et al. \cite{Lee2009} proposed a method based on fractional Fourier transform (FrFT) to extract energy from the fingerprint in the spectrum image. Energy differences of live (high energy)  and fake (low energy) fingerprints are used as an indicator for fake fingerprint detection. The major limitation of single feature based approaches is that, it fails to perform equally over different fingerprint sensors and materials.

In \cite{Choi2009}, Choi et al. proposed a novel fingerprint liveness detection method using static features such as histogram, directional contrast, ridge thickness, and ridge signal of each fingerprint image. The extracted features are fused and fed to support vector machine for live or fake fingerprint classification over the custom-made database. Nikam et al. \cite{Nikam2008} proposed  texture and wavelet-based fingerprint liveness detection method using structural, orientation, roughness, smoothness and regularity differences of diverse regions in a fingerprint image. They have utilized local binary pattern (LBP) for texture analysis and wavelet energy features for ridge frequency and orientation feature estimation. Ghiani et al. \cite{Ghiani2013} proposed Binarized Statistical Image Features (BSIF) similar to Local Binary Pattern
and Local Phase Quantization for fingerprint texture based liveness detection. In \cite{Galbally2009}, Galbally et al. proposed fingerprint liveness detection using quality-related features. They have considered ridge strength measures, ridge continuity measures and ridge clarity measures for liveness detection over LivDet 2009 database. An improved study of this work is proposed in \cite{GALBALLY2012}, which provides the liveness detection results on multi-scenario datasets of LivDet 2009 and ATVS. The proposed method provides a more robust solution for entirely diverse testing scenarios. In \cite{Galbally2014}, Galbally et al. have also proposed an image quality assessment based method for fake iris, fingerprint and face detection. The proposed approach considers 25 general image quality features (mean squared error, signal to noise ratio, structural content, etc.) extracted from a single image to distinguish between real and fake samples in multiple biometric systems. Multiple features extracted in these methods are not able to perform well over different types of sensors.

In \cite{Ghiani2012}, Ghiani et al. experimented with several state-of-the-art fingerprint liveness detection algorithms on the benchmark datasets available at LivDet 2011. The results exhibit that LBP based approach \cite{maltoni2009handbook} is most effective over four datasets used in LiveDet 2011 competition. Huang et al. \cite{HUANG2015}, proposed a study on evaluation of fake fingerprint databases utilizing support vector machine (SVM) classification algorithm. In their study, three  public fake fingerprint databases (LivDet 2013, ATVS, and CASIA) are evaluated by comparing the classification accuracies of SVM classifier with different feature vector (spatial features, detailed ridge features, and Fourier spectrum features). Their study shows latex and body doubles fabricated fingerprints are most difficult to discriminate. Xia et al. \cite{Xia2017} proposed fingerprint liveness detection using elements of co-occurrence array obtained from image gradients. A brief review of LivDet datasets (2009, 2011, 2013, 2015) and various algorithms submitted in these competitions can be found in \cite{GHIANI2017}.

Software-based liveness detection approaches relying only on one impression results in faster detection of fake or real fingerprints. However, none of the existing approaches are able to classify fake and live fingers with acceptable error rates so far. They usually exploit a limited set of features to capture the vitality by exploiting different aspects of the fingerprint.
\vspace{-5mm}
\section{Proposed method}
\begin{figure}[h]
	\vspace{-4mm}
	\centering
	\resizebox{\textwidth}{!}{
		\includegraphics[]{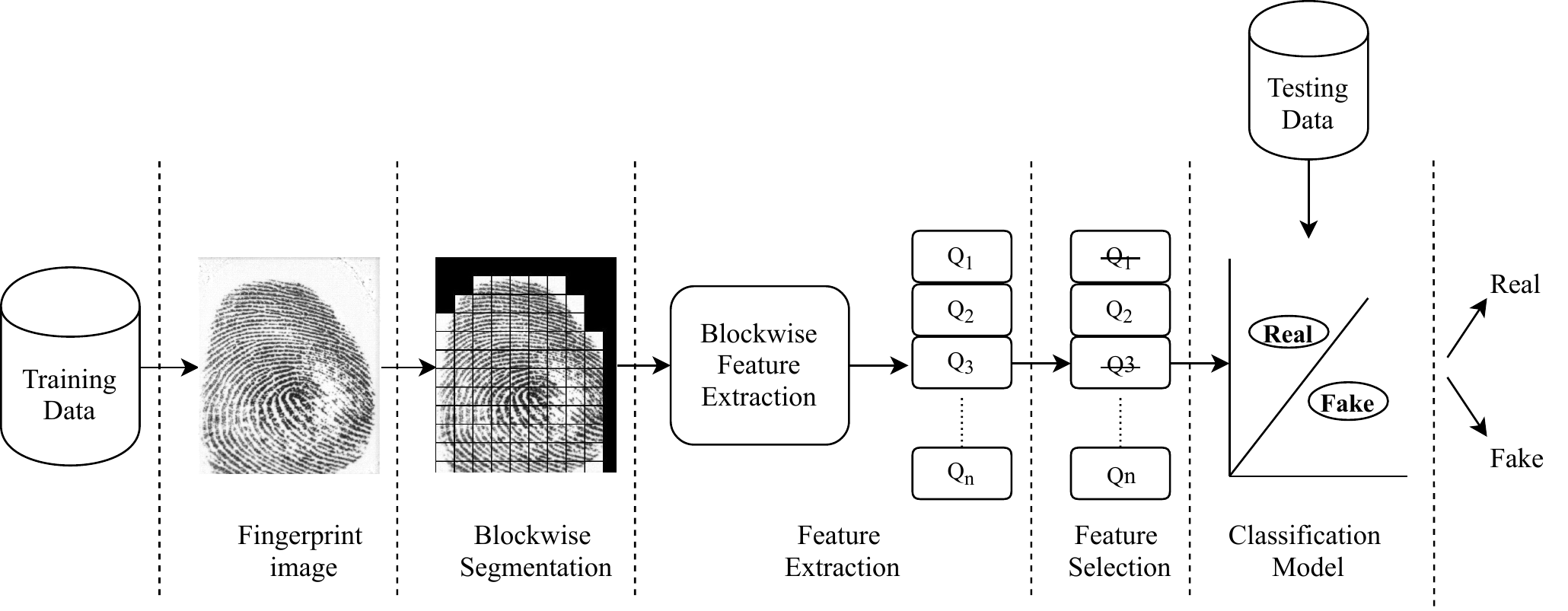}}
	\caption{Schematic diagram of the proposed quality based liveness detection method for each sensor}
	\vspace{-5mm}
	\label{BlockDiagram}
\end{figure}

\vspace{-2mm}
In order to reduce overall error rates of liveness detection system, novel quality features are proposed and eventually integrated with existing ones to find the optimal set of features. A combination of the proposed and existing features is expected to achieve better performance. Further, we propose a feature selection process that selects best feature subset for each sensor. The schematic diagram of the proposed approach is illustrated in Fig. \ref{BlockDiagram}.

 \vspace{-5mm}

\subsection{Feature extraction}
\begin{figure}[b]
	\centering	
	\includegraphics[width=12cm,height=10cm,keepaspectratio]{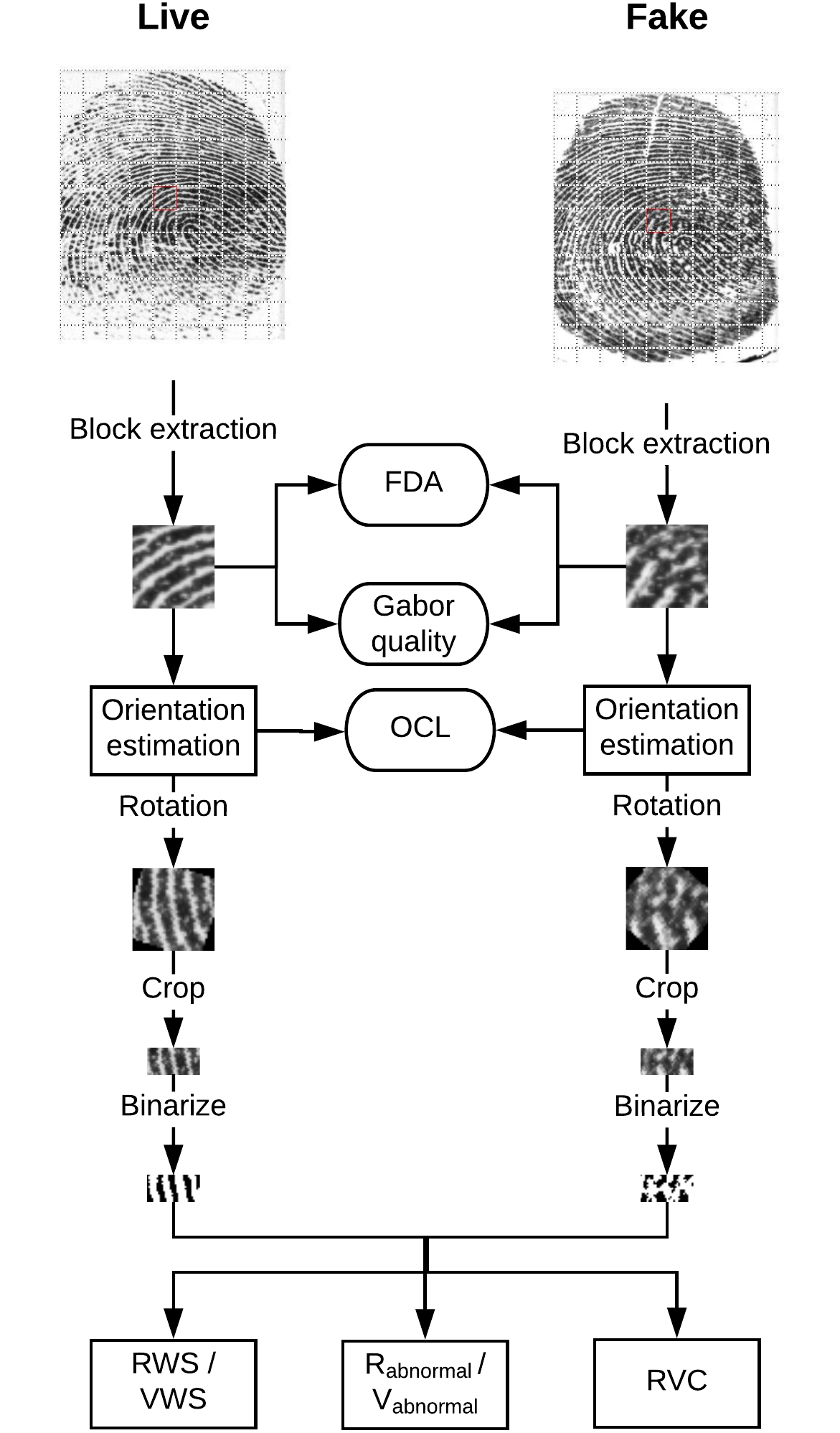}
	\caption{Graphical overview of feature extraction from a live and fake fingerprint image}	
	\vspace{-4mm}
	\label{Fextract}
\end{figure}

\vspace{-2mm}
Image quality features based on ridge-valley properties of fingerprints are vital to detect fake fingerprints. The elasticity of the materials used for fabricating replica of fake fingerprints introduces non-uniformity in the ridge-valley structure of the captured image. As ridges and valleys are core part of a fingerprint image, scanning the differences between the ridge-valley structure of real and fake fingerprints is crucial in fingerprint liveness detection. Characteristics of the ridge-valley structure of real and fake fingerprints are given in Fig. \ref{RVchar}. Based on these minute observation, we propose RWS, VWS, $R_{ab}$, $V_{ab}$, and RVC features. In addition to these features, FDA, OCL, and G features are also considered for liveness detection in this work. All the features are extracted on the local level by rotating the block of a fingerprint image using orientation estimation method adapted from \cite{tabassi05} to make ridge-valley structure vertical. Further, various features are extracted from these blocks of real and fake fingerprint images. An overview of the extraction of different features is given in Fig. \ref{Fextract}.  Details of the features extracted in this work is described as follows: 

\begin{figure}[t]
	\vspace{-2mm}
	\centering	
	\includegraphics[width=10cm,height=10cm,keepaspectratio]{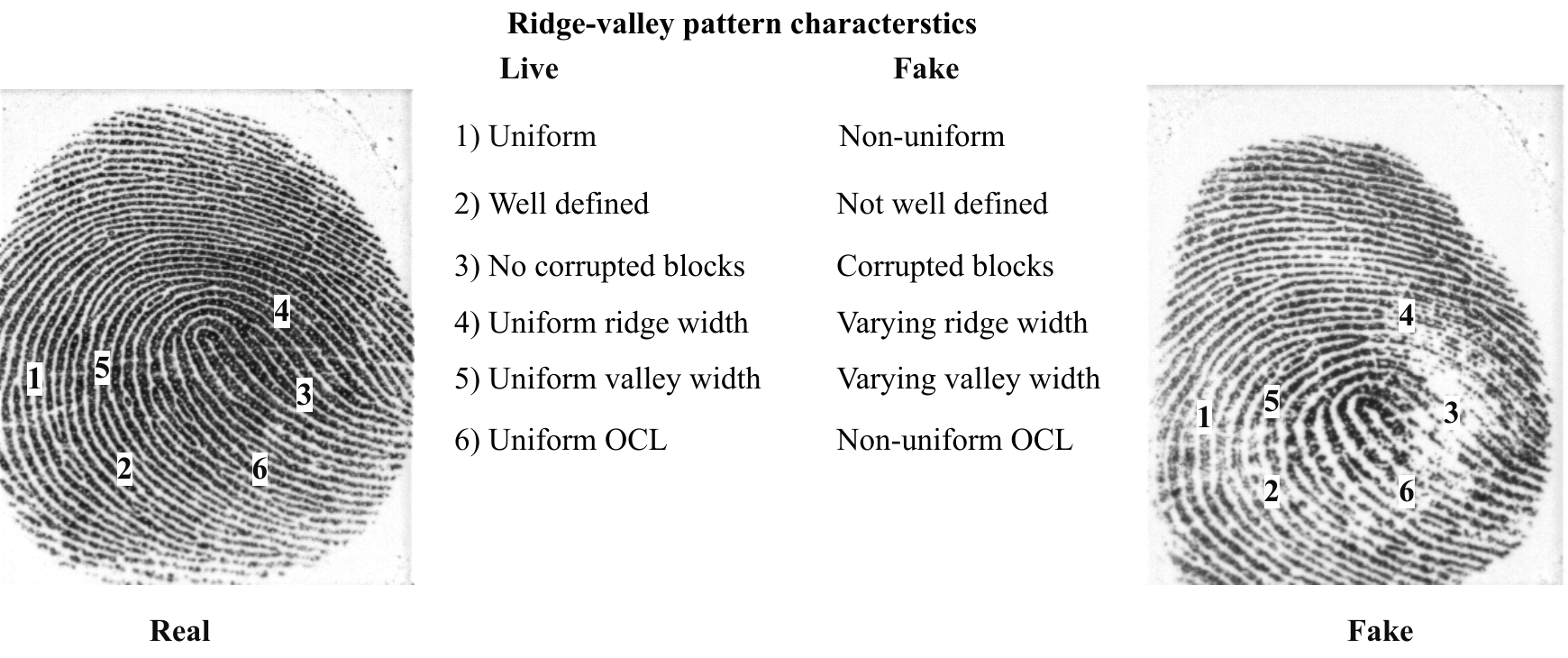}
	\caption{Ridge-valley characteristics of the live and fake fingerprint image}	
	\vspace{-5mm}
	\label{RVchar}
\end{figure}

\vspace{-3mm}
\begin{itemize}
	\item \bf{ Ridge width smoothness/Valley width smoothness:} \normalfont  It indicates smoothness of ridge and valley width in different blocks of a fingerprint image. As seen from Figure \ref{RVchar}, real fingerprints have a near constant ridge and valley width while the fake fingerprints have varying ridge and valley width due to the elasticity of the fabrication materials and non-uniform pressure while fabricating fake fingerprints. Ridge and valley width smoothness is measured by first cropping the block having a vertical ridge-valley structure to remove invalid regions. The resulting block is binarized using linear regression. Thereafter, the width of ridges and valleys is computed for each horizontal line of the block having alternate ridge-valley structure. Widths of the each ridge and valley of the block will be aligned. Finally, $RWS^{l}$ and $VWS^{l}$ of each block is computed by averaging the standard deviation of each ridge and each valley widths across different horizontal lines using Eq. \ref{RWS} and \ref{VwidthS}. An example of the RWS map and VWS map is depicted in Fig. \ref{RVS}.
\vspace{-2mm}	
	\begin{align}
	\label{RWS}
	RWS^{l}=\frac{1}{|R|}\sum\limits_{c=1}^{c=|R|} \sqrt{\frac{1}{n}\sum\limits_{r=1}^{r=n}(rw_{rc} - \overline{rw_c})^2}  \hspace{4mm}  \\ \hspace{5mm} where \hspace{2mm}	rw = \begin{pmatrix}
	{rw_{11}}& {rw_{12}} & \hdots  {rw_{1c}}\\
	{rw_{21}} &{rw_{22}} & \hdots {rw_{2c}}\\
	\vdots & \vdots & \vdots\\
	{rw_{r1}}& {rw_{r2}}& \hdots {rw_{rc}}\\
	\end{pmatrix} and \hspace{3mm} \overline{rw_c}=\frac{1}{n}\sum\limits_{r=1}^{r=n}rw_{rc} \notag    
	\end{align}
	
	\begin{align}
	\label{VwidthS}
	VWS^{l}=\frac{1}{|V|}\sum\limits_{c=1}^{c=|V|} \sqrt{\frac{1}{n}\sum\limits_{r=1}^{r=n}(vw_{rc} - \overline{vw_c})^2} \hspace{4mm}\\ \hspace{5mm} where \hspace{2mm}	vw = \begin{pmatrix}
	{vw_{11}}& {vw_{12}} & \hdots  {vw_{1c}}\\
	{vw_{21}} &{vw_{22}} & \hdots {vw_{2c}}\\
	\vdots & \vdots & \vdots\\
	{vw_{r1}}& {vw_{r2}}& \hdots {vw_{rc}}\\
	\end{pmatrix} and \hspace{3mm} \overline{vw_c}=\frac{1}{n}\sum\limits_{r=1}^{r=n}vw_{rc}\notag
	\end{align}

	here, $n$ is number of rows, $|R|$ and $|V|$ are number of ridges and valleys in the block and $rw$ and $vw$ are the matrix containing widths of the ridges and valleys across $n$ rows of a block, respectively. $\overline{rw_c}$ and $\overline{vw_c}$ are mean of the widths of a single ridge and valley stored in each column $c=1 \hdots c$ of $rw$ and $vw$, respectively.
	
	\begin{figure}[t]
	\centering
	
	\subfigure[]
	{
		\includegraphics[width=0.7in]{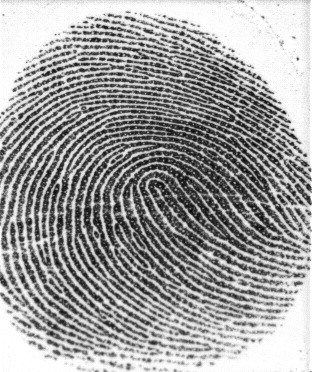}
		\includegraphics[width=0.7in]{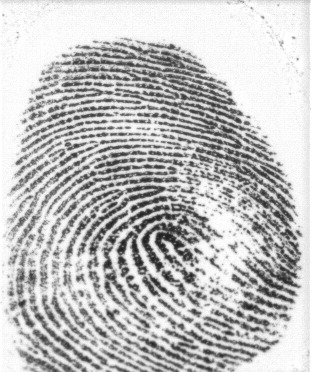}
		\label{Qap dry}
	}
	\hspace{0.02mm}
	\subfigure[]
	{
		\includegraphics[width=0.7in]{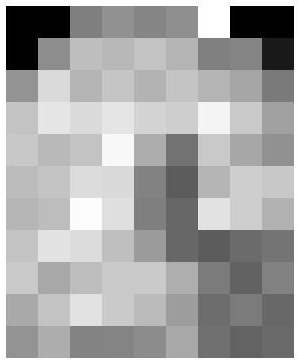}
		\includegraphics[width=0.7in]{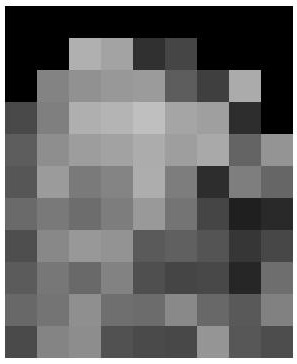}
		\label{Qap nd}
	}		
	\subfigure[]
	{
		\includegraphics[width=0.7in]{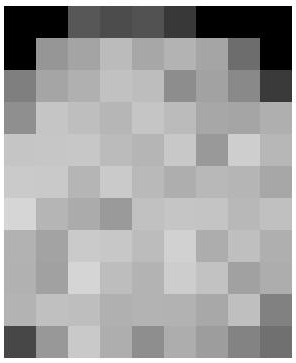}
		\includegraphics[width=0.7in]{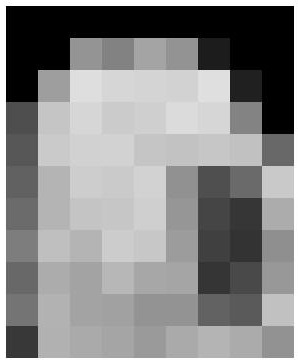}
		\label{Qap nd}
	}		
	
	\caption{Computation of the RWS/VWS for live and fake fingerprints. (a) real and fake fingerprints (b) RWS of live and fake fingerprints (c) VWS of live and fake fingerprints}
	\vspace{-5mm}
	\label{RVS}
\end{figure}

	\item \bf{Number of abnormal ridge/valley:} \normalfont Generally, a 500 dpi fingerprint image contains 5-10 pixel wide ridges and valleys \cite{maltoni2009handbook}. Some of the blocks of fake fingerprints exhibit an abnormal change in the ridge width due to the elasticity of the material used for fake fingerprint fabrication as seen in Fig. \ref{RVchar}. A ridge or valley in a local block is considered as abnormal if the deviation of its widths in different rows of the block is above a threshold width deviation $t_w =1.03$. The threshold value is obtained by allowing maximum width change of 2 pixels between any two consecutive horizontal rows of the block for a particular ridge or valley. $R_{ab}^{l}$ and $V_{ab}^{l}$ in a local block is computed using Eq. \ref{Rab} and Eq. \ref{Vab}.
	
	\begin{equation}
	R_{ab}^{l}=\sum\limits_{c=1}^{c=|R|}\begin{cases}
	1, & \text{if $std(rw_c)>t_w$}.\\
	0, & \text{otherwise}.
	\end{cases}
	\label{Rab}
	\end{equation}
	\begin{equation}
	V_{ab}^{l}=\sum\limits_{c=1}^{c=|V|}\begin{cases}
	1, & \text{if $std(vw_c)>t_w$}.\\
	0, & \text{otherwise}.
	\end{cases}
	\label{Vab}
	\end{equation}

	\item \bf{Ridge valley clarity:} \normalfont Separation between two consecutive ridges and valleys in a local block of the live fingerprint image is almost constant. On the other hand, this separation can be varying in fake fingerprints due to the varying widths of ridges and valleys in a block. To measure ridge-valley clarity, average ridge and valley width of a block is computed. The number of misclassified ridge pixels in the valley region between the two consecutive ridges and  the number of misclassified valley pixels in the ridge region between the two consecutive valleys are counted. $RVC^{l}$ map for local blocks of real and fake fingerprint image is shown in Fig. \ref{RVC}. 
	\begin{equation}
	RVC^{l}=\frac{(rw - \overline{rw})+(vw - \overline{vw})}{rw_{sum}+vw_{sum}}
	\label{RVclarity}
	\end{equation}	
	
	here, $rw$ and $vw$ contain the widths of ridges and valleys in different rows of a vertically rotated block given in Eq. 1 and Eq. 2 and $\overline{rw}$ and $\overline{vw}$ is the average width of ridge and valley in the block. $rw_{sum}+vw_{sum}$ indicates total number of ridge and valley pixels in the block.
	\begin{figure}[t]
		\centering
		
		\subfigure[]
		{
			\includegraphics[width=0.7in]{Real41_13_2.jpg}
			\includegraphics[width=0.7in]{Fake41_13_2.jpg}
			\label{Qap dry}
		}
		\hspace{0.02mm}
		\subfigure[]
		{
			\includegraphics[width=0.7in]{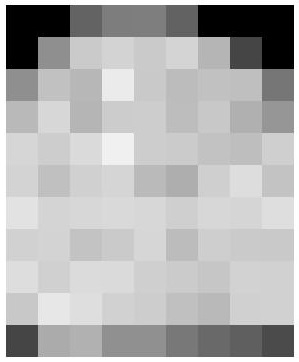}
			\includegraphics[width=0.7in]{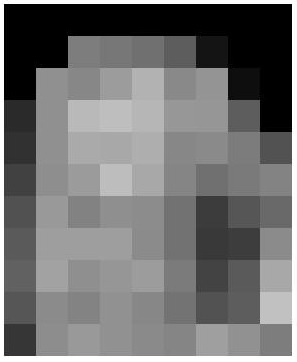}
			\label{Qap nd}
		}

		\caption{Computation of the RVC for live and fake fingerprints. (a) live and fake fingerprints (b) RVC map of live and fake fingerprints}
		\vspace{-5mm}
		\label{RVC}
	\end{figure}
	
	\item \bf{Frequency domain analysis (FDA):} \normalfont FDA \cite{olsen2016} of a local block is computed by extracting 1D signature of ridge-valley structure. DFT of this 1D signature is computed to obtain the frequency of the sinusoidal ridge-valley structure. Live fingerprint have uniform frequency of sinusoidal ridge-valley structure while it varies in fake fingerprints. The local FDA quality ($FDA^{l}$) is computed as follows:
	
	\begin{center}
		$FDA^{l}=\frac{A(F_{max})+C(A(F_{max}-1))+A(F_{max}+1)}{\sum_{F=1}^{F=N/2}A(F)}$          \label{RVclarity}
	\end{center}	
	
	here, C = 0.3 as per definition appearing in ISO/IEC TR29794-4:2010. F and A represents frequency and amplitude,respectively. Constant is used to retain an attenuated amplitude of the frequency bands immediately surrounding $F_{max}$. The value of $FDA^{l}$ is set to 1 if $F{max} = 1$ or $F{max} = A(end)$ as
	both A(0) and A(end + 1) are not accessing valid indices.
	
		\begin{figure}[t]
		\centering
		
		\subfigure[]
		{
			\includegraphics[width=0.7in]{Real41_13_2.jpg}
			\includegraphics[width=0.7in]{Fake41_13_2.jpg}
			\label{Qap dry}
		}
		\hspace{0.02mm}
		\subfigure[]
		{
			\includegraphics[width=0.7in]{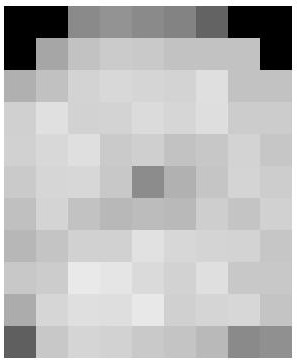}
			\includegraphics[width=0.7in]{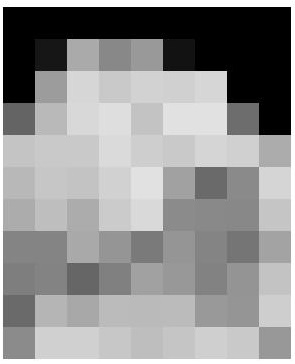}
			\label{Qap nd}
		}		
		
		\caption{Computation of the OCL for live and fake fingerprints. (a) live and fake fingerprints (b) OCL of live and fake fingerprints}
	\vspace{-5mm}
		\label{OCL}
	\end{figure}

	\item \bf{Orientation certainty level (OCL):} \normalfont OCL \cite{olsen2016} is measured using the intensity gradient in a local block where the energy concentration along the dominant direction of ridges is estimated. OCL is computed as the ratio of the two eigenvalues of the covariance matrix computed using the gradient vector. The covariance matrix ($C$) \cite{tabassi05} of the block ($m \times n$) using intensity gradient with centered differences method is computed as follows:

	\begin{center}
		$C$=$\frac{1}{m*n}\sum_{m*n}^{}\begin{Bmatrix}
		\begin{bmatrix}
		dx \\
		dy
		\end{bmatrix}
		\begin{bmatrix}
		dx & dy
		\end{bmatrix}
		\end{Bmatrix}$=$\begin{bmatrix}
		a & c\\
		c & b
		\end{bmatrix}$
	\end{center}
	
	From the covariance matrix C the eigenvalues
	($\lambda_{min},\lambda_{max}$) are computed as:
	\begin{center}
		$
		\lambda_{min}=\frac{a+b-\sqrt{(a-b)^2+4c^2}}{2}	
		$
	\end{center} 
	\begin{center}
		$
		\lambda_{max}=\frac{a+b+\sqrt{(a-b)^2+4c^2}}{2}
		$
	\end{center}
	These eigenvalues yields the $OCL^{l}$ of a block as follows:
	
	\begin{center}
		$
		OCL^{l}=\begin{cases}
		1 - \frac{\lambda_{min}}{\lambda_{max}}, & \text{if $\lambda_{max}>0$}\\
		0, & \text{otherwise}
		\end{cases}
		$
		\label{Vabnormal}
	\end{center}
	
OCL map for real and fake fingerprint image is shown in Fig. \ref{OCL}.

	\item \bf{Gabor quality:} \normalfont Gabor filter is used to measures the quality of the local blocks of real and fake fingerprint images \cite{olsen2016}. Gabor filter bank operates on per-pixel basis by calculating the standard deviation of the Gabor filter bank responses. The strength of the response for blocks with regular ridge–valley pattern will be high for one or a few filters having neighborhood orientations. In the blocks containing unclear ridge-valley structure, the Gabor response of all orientations will be low and constant. The standard	deviation of the Gabor filter bank responses indicates the Gabor quality (G) of the block. Fig. \ref{gabor} shows the Gabor quality map of real and fake fingerprint images.
	
	\begin{figure}[t]
		\centering
		
		\subfigure[]
		{
			\includegraphics[width=0.7in]{Real41_13_2.jpg}
			\includegraphics[width=0.7in]{Fake41_13_2.jpg}
			\label{Qap dry}
		}
		\hspace{0.02mm}
		\subfigure[]
		{
			\includegraphics[width=0.7in]{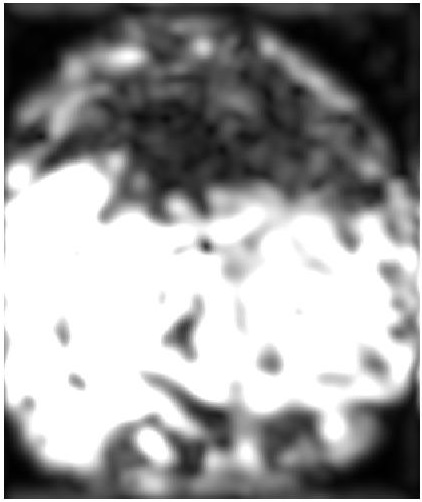}
			\includegraphics[width=0.7in]{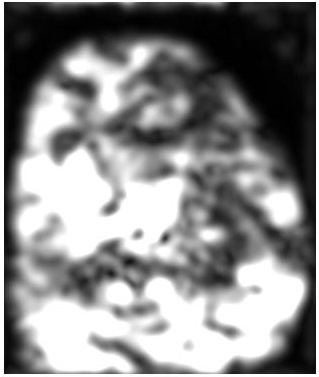}
			\label{Qap nd}
		}		
		
		\caption{Computation of the Gabor quality for live and fake fingerprints. (a) live and fake fingerprints (b) Gabor quality of live and fake fingerprints}
			\vspace{-5mm}
		\label{gabor}
	\end{figure}

\end{itemize}
	\vspace{-6mm}
\subsubsection{Quality vectors from the local qualities}\label{featurevector}
	\vspace{-2mm}
\begin{itemize}
	\item \bf{Mean of local quality features:} \normalfont The mean feature value ($Q^{\mu}$) of the $N \times M$ local quality features ($Q^{l}$) is computed using Eq. \ref{Mean}.
	
	\begin{equation}
	Q^{\mu}=\frac{1}{N \times M}\sum\limits_{i=1}^{i=N}\sum\limits_{j=1}^{j=M}Q^{l} \hspace*{0.3mm}   
	\label{Mean}
	\end{equation}

	here, Q is one of the RWS, VWS, $R_{ab}$, $V_{ab}$, RVC, FDA, OCL, and G.
	
	\item  \bf{Standard deviation of local quality features:} \normalfont The standard deviation value ($Q^{\sigma}$) of the $N \times M$ local quality values ($Q^{l}$) is computed using Eq. \ref{sd}.
	
	\begin{equation}
	Q^{\sigma}=\Bigg(\frac{1}{N \times M - 1}\sum\limits_{i=1}^{i=N}\sum\limits_{j=1}^{j=M}\bigg(Q^{l} - Q^{\mu}\bigg)^2 \Bigg)^\frac{1}{2} \hspace*{0.3mm}   
	\label{sd}
	\end{equation}	
	
	here, Q is one of the RWS, VWS, RVC, FDA, and OCL.
\end{itemize}

The feature vector is produced as a concatenation of individual quality features as:

\hspace*{20mm}$ Q= \bigg\{ {RWS}^{\mu}, {VWS}^{\mu}, {RVC}^{\mu}, {FDA}^{\mu}, {OCL}^{\mu},\\\ \hspace*{36mm} {RWS}^{\sigma}, {VWS}^{\sigma}, {RVC}^{\sigma}, {FDA}^{\sigma}, {OCL}^{\sigma},\\\ \hspace*{36mm} R_{ab}^{\mu}, V_{ab}^{\mu}, G^{\mu}    \bigg\} $

\vspace{-2mm}
\subsection{Feature selection}
\vspace{-2mm}
Higher dimensionality of the feature set may be a curse to the classification results. It is possible that the best classifying results are not obtained utilizing all the proposed features. Additionally, the time required to perform classification (extracting all features) is a fundamental parameter which influences the performance of the classification problem. Therefore, a feature selection phase is required to identify the best-performing feature set and subsequently reducing the time required for feature extraction. The optimal feature subset resulting in highest classification accuracy for each sensor is selected using Sequential Forward Floating Selection (SFFS) technique \cite{PUDIL1994}. The SFFS algorithm determines the best subsets of features with highest discriminating capability than other for each sensor. The SFFS method is deterministic single solution feature selection algorithm proposed in \cite{PUDIL1994} having remarkable performance over other feature selection schemes \cite{Jain1997}. The selected optimal feature subset is used for validating the classification results on the testing set of each sensor.

\begin{figure}[t]
	\centering	
	
	\subfigure[]
	{
		
		\includegraphics[width=0.7in]{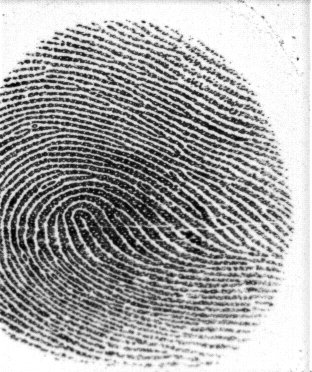}
		\includegraphics[width=0.7in]{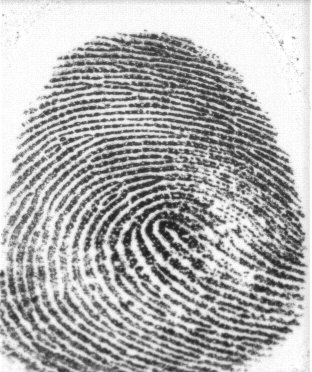}
		
	}
	
	\subfigure[]
	{
		
		\includegraphics[width=0.6in]{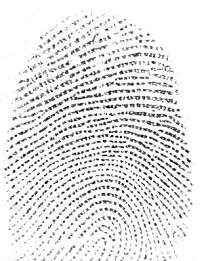}
		\includegraphics[width=0.6in]{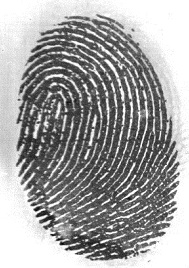}
		\includegraphics[width=0.6in]{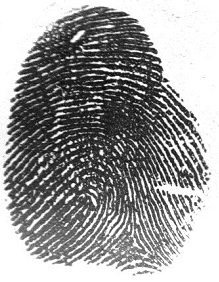}
		\includegraphics[width=0.6in]{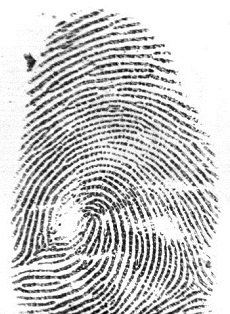}
		
	}
	
	\subfigure[]
	{	
		\includegraphics[width=0.6in]{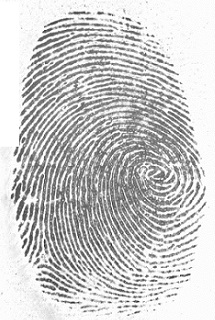}
		\includegraphics[width=0.6in]{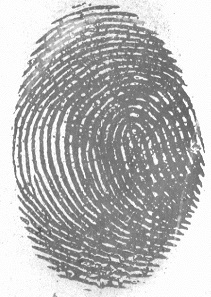}
		\includegraphics[width=0.6in]{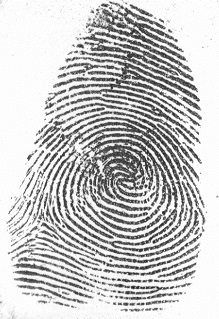}
		\includegraphics[width=0.6in]{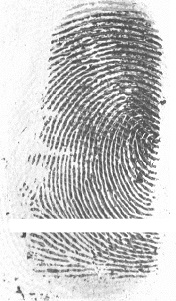}

	}
	
	\caption{Examples of real and fake fingerprint images obtained from LiveDet 2009 datasets. (a) Biometrika; real, and silicone (left to right)(b) Crossmatch; real, gelatin, playdoh, and silicone (left to right) (c) Identix;  real, gelatin, playdoh, and silicone (left to right) }
	\label{sampleimages}
	\vspace{-4mm}
\end{figure}

	\vspace{-5mm}
\section{Experimental results}
	\vspace{-2mm}
The performance of the proposed liveness detection method is evaluated on LivDet 2009 \cite{LivDet09} and LivDet 2015 \cite{LivDet15} databases. The general distribution of the fingerprint images between train and test set of LivDet 2009 and 2015 is given in Table \ref{LiveDet200915}. The datasets in both of the LivDet competition is divided into training set and testing set for different sensors. There is no overlapping between train and test set to obtain totally unbiased results. Best feature subset is selected using training set and used for modeling a classification system while evaluation results are obtained on test set.
	\vspace{-6mm}
\subsection{Dataset and performance metrics}
	\vspace{-2mm}
LivDet 2009 database is composed of three datasets containing live and fake fingerprint images. Each dataset is captured with a different flat optical sensor: (i) Biometrika FX2000, (ii) CrossMatch Verifier 300CL, and (iii) Identix DFR2100 composed of two subsets, one for training and the other one for testing the results of trained liveness detection algorithm. The fake fingerprints are generated using three different materials: silicone, gelatin, and playdoh with the cooperation of the user. Some examples of the live and fake fingerprints found in LivDet 2009 datasets is given in Fig. \ref{sampleimages}. In LivDet 2015 database, there are 4 datasets captured with (i) GreenBit (ii) Biometrika (iii) Digital persona (iv) CrossMatch sensors. The fake images in the train set are fabricated using ecoflex, gelatin, latex, wood glue, playdoh, and body double materials. The testing set includes spoofs fabricated using new materials, which are not part of the training set. These new materials included liquid ecoflex and RTV for Biometrika, Digital Persona, and Green Bit sensors, and OOMOO and gelatin for Crossmatch sensor.

\begin{table}[t]
	\centering
	\caption {Detailed description of the LivDet 2009 and LivDet 2015 datasets used in the experiments}

	\label{LiveDet200915}
	\resizebox{\textwidth}{!}{
		\begin{tabular}{@{}llllll@{}}
			\toprule
			\textbf{Dataset}     & \textbf{Sensor} & \textbf{\begin{tabular}[c]{@{}l@{}}Resolution\\ (dpi)\end{tabular} } & \textbf{\begin{tabular}[c]{@{}l@{}}Live Images\\ Train / Test\end{tabular}} & \textbf{\begin{tabular}[c]{@{}l@{}}Fake Images\\ Train / Test\end{tabular}} & \textbf{\begin{tabular}[c]{@{}l@{}}Spoof \\ Materials\end{tabular}}                                  \\ \midrule
			\textbf{LivDet 2009} & Biometrika      & 569                 & 520/1480                                                                    & 520/1473                                                                    & Silicone                                                                                             \\\\
			\textbf{}            & CrossMatch      & 500                 & 1000/3000                                                                   & 1000/3000                                                                   & Gelatin, PlayDoh, Silicone                                                                           \\\\
			\textbf{}            & Identix         & 686                 & 750/2250                                                                    & 750/2250                                                                    & Gelatin, PlayDoh, Silicone                                                                           \\\\
			\textbf{LivDet 2015} & Biometrika      & 1000                & 1000/1000                                                                   & 1000/1500                                                                   & \begin{tabular}[c]{@{}l@{}}Ecoflex, Gelatin, Latex, \\ WoodGlue, Liquid Ecoflex, \\ RTV\end{tabular} \\\\
			& CrossMatch      & 500                 & 1000/1000                                                                   & 1473/1448                                                                   & \begin{tabular}[c]{@{}l@{}}BodyDouble, Ecoflex,\\ PlayDoh, OOMOO,\\ Gelatin\end{tabular}          \\   \\
			& Digital Persona & 500                 & 1000/1000                                                                   & 1000/1500                                                                   & \begin{tabular}[c]{@{}l@{}}Ecoflex, Gelatin, Latex, \\ WoodGlue, Liquid Ecoflex, \\ RTV\end{tabular} \\\\
			& GreenBit        & 500                 & 1000/1000                                                                   & 1000/1500                                                                   & \begin{tabular}[c]{@{}l@{}}Ecoflex, Gelatin, Latex, \\ WoodGlue, Liquid Ecoflex, \\ RTV\end{tabular} \\ \bottomrule
	\end{tabular}}
	\vspace{-2mm}
\end{table}
The classification performance of the proposed system is evaluated adopting the same parameters used for LivDet competitions \cite{LivDet09,LivDet15}. The threshold score for determining liveness in a fingerprint is set at 0.5.  The fingerprint image with score more than 0.5 is considered as the real one, while it is fake if the value is less than or equal to 0.5. In particular, parameters which are computed based on this threshold are:

\begin{itemize}
	\item Ferrlive: It represents rate of live fingerprints misclassified as fake.
	\item Ferrfake: It represents rate of fake fingerprints misclassified as live.
	\item  Average classification error (ACE) : The overall performance indicator of the system is given by ACE defined in Eq. \ref{ace}:
	
	\begin{equation}
	ACE = \frac{ Ferrlive + Ferrfake}{2}   
	\label{ace}
	\end{equation}
\end{itemize}
	\vspace{-4mm}
\subsection{Random forest classifier}
	\vspace{-2mm}
Random forest is utilized to train the classifier with the proposed feature vector. Random Forest combines multiple decision trees, minimizing the variance that comes with single complex trees. The chance of stumbling around a classifier that doesn't perform well because of variance between train and test set features is reduced while using multiple decision trees in the random forest. The risk of overfitting is also reduced by averaging the classification results of several trees.

	\vspace{-4mm}
\subsubsection{Training the classifiers}
	\vspace{-2mm}
The training set each of the different acquisition devices of LivDet 2009 is used for finding the optimal feature subset. The best feature subset found using SFS technique for each of the sensors are reported in Table \ref{featureSelection}, where \checkmark  means that the feature is included in the optimal subset. Results shown in Table \ref{featureSelection} indicates that except the $G^\mu$ feature all the features are included in at least one of the optimal subsets. It also indicates that most of the proposed features  are appropriate for fingerprint liveness detection. Features $RWS^\sigma$, and $R_{ab}^\mu$ indicating higher discriminating capabilities than other features as these features are included in optimal features of all three datasets. $RWS^\mu$, $VWS^\mu$, $V_{ab}^\mu$, and $FDA^\mu$ provides good discriminative capabilities as these features are included in optimal features of two subsets. On the other hand, least useful features are $VWS^\sigma$, $RVC^\mu$, $RVC^\sigma$, $FDA^\sigma$, $OCL^\mu$, and $OCL^\sigma$. These features are included only in one sensor's optimal feature subset. The ACE using optimal feature subset of each of the dataset is validated on the test set. 

\begin{table}[t]
	\centering
	\caption{Optimal subset of features for the different datasets provided in LivDet 2009 database. The symbol \checkmark means that the feature is included in the optimal feature subset}
	\vspace{2mm}
	\label{featureSelection}
	\begin{tabular}{lllll}
		\hline
		Number & Features     & Biometrika & CrossMatch & Identix \\\hline
		1 &	$RWS^\mu$    &  &   \checkmark          &  \checkmark        \\
		2 &	$RWS^\sigma$ &   \checkmark          &   \checkmark          &  \checkmark        \\
		3 &	$VWS^\mu$    &  \checkmark           &             & \checkmark         \\
		4 &	$VWS^\sigma$ &            & \checkmark           &         \\
		5 &	$R_{ab}^\mu$ &  \checkmark           &  \checkmark           &  \checkmark        \\
		6 &	$V_{ab}^\mu$ &            &  \checkmark           & \checkmark         \\
		7 &	$RVC^\mu$    &     \checkmark        &            &         \\
		8 &	$RVC^\sigma$ &            &   \checkmark         &         \\
		9 &	$FDA^\mu$    &   \checkmark          &             &   \checkmark       \\
		10 &	$FDA^\sigma$ &     \checkmark        &            &         \\
		11 &	$OCL^\mu$    &            &            &      \checkmark    \\
		12 &	$OCL^\sigma$ &  \checkmark           &            &         \\
		13 &	$G^\mu$      &            &            &         \\
		\hline
	
	\end{tabular}
\vspace{-4mm}
\end{table}
\begin{table}[b]
	\centering
	\caption{Performance results in terms of Ferrfake, Ferrlive, and ACE (in \%) of the proposed liveness detection method on LivDet 2009 datasets}
	\vspace{2mm}
	\label{LivDet09Results}
	\begin{tabular}{llll}
		\hline
		Sensors    & Ferrlive & Ferrfake & ACE \\\hline
		Biometrika   &    6.8   &  \begin{tabular}[t]{@{}l@{}} 10.2 \hspace{6.5mm}     Silicone\end{tabular}         &  8.5  \\
		Crossmatch    &    4.7  & \begin{tabular}[t]{@{}l@{}}  0.0 \hspace{8mm}     Gelatin\\     0.6        \hspace{8mm}   PlayDoh\\        18.9        \hspace{6.5mm} Silicone  \end{tabular}         &  5.6   \\
		Identix      &   1.2      &  \begin{tabular}[t]{@{}l@{}}  0.1\hspace{9mm}     Gelatin\\    0.8         \hspace{8mm}   PlayDoh\\          7.5      \hspace{8mm} Silicone\end{tabular}        &  2.0   \\
		\textbf{Average}    &        &          &  5.3  \\\hline
	\end{tabular}
	\vspace{-8mm}
\end{table}

	\vspace{-3mm}
\subsubsection{Validation}
\begin{figure}[t]
	\centering
	\resizebox{0.7\textwidth}{!}{
		\includegraphics[]{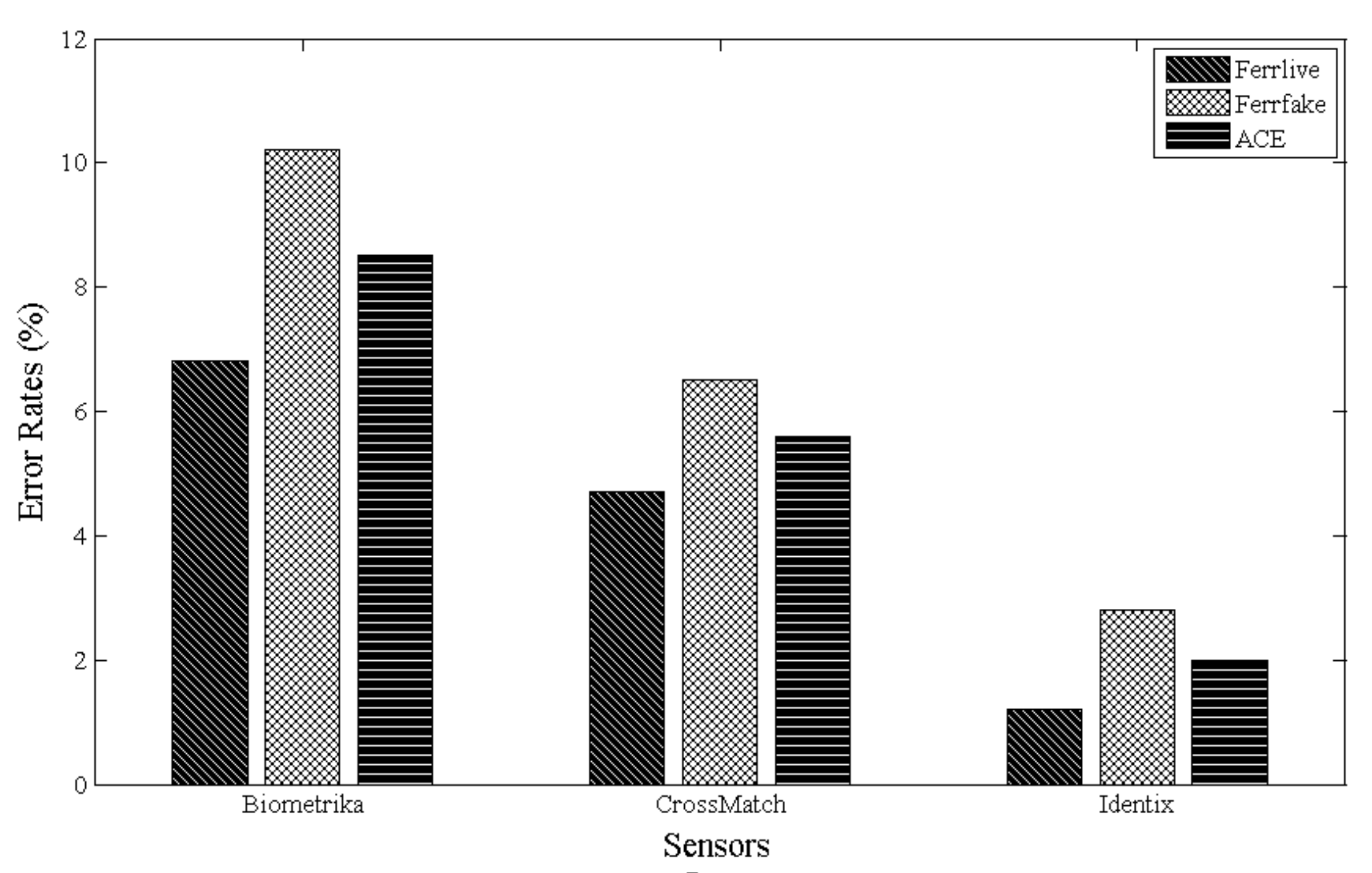}}
	\caption{Error rates (Ferrlive, Ferrfake, and ACE) for the Biometrika, CrossMatch, and Identix sensors}
	\vspace{-4mm}
	\label{ErrorRates}
\end{figure}

	\vspace{-2mm}
The performance of the proposed liveness detection is evaluated using the best feature subset found on the training sets of each dataset. The ACE for testing set of each dataset is given in Table \ref{LivDet09Results} with the overall ACE of the system.Results are analyzed in terms of Ferrlive and Ferrfake for each fabrication material (gelatin, playdoh, and silicone). It is observed that the fake fingerprints fabricated using silicone are most difficult to detect because of higher similarity with the live fingerprints. Error rates for each of the sensor are plotted in Fig. \ref{ErrorRates}. 
It can be seen that Ferrlive, Ferrfake, and ACE for Identix sensor is least followed by CrossMatch and Biometrika sensor. The overall ACE for the LivDet 2009 database is 5.3\%.

\begin{figure}[t]
	\vspace{-2mm}
	\centering	
	\resizebox{0.49\textwidth}{!}{
		\subfigure[]
		{ 		
			\includegraphics[width=18cm, height=14cm]{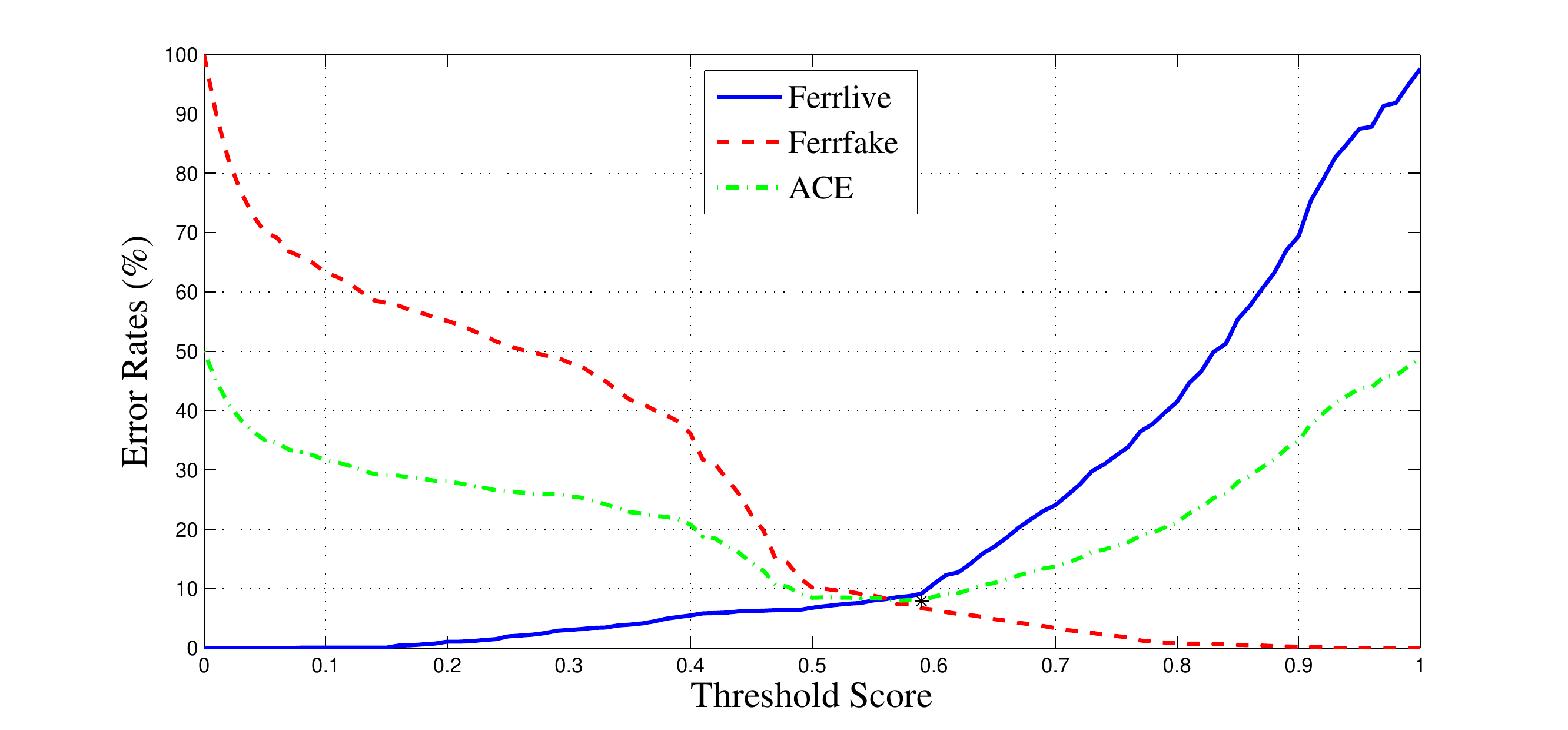}		
	}}
	\resizebox{0.49\textwidth}{!}{
		\subfigure[]
		{ 		
			\includegraphics[width=18cm, height=14cm]{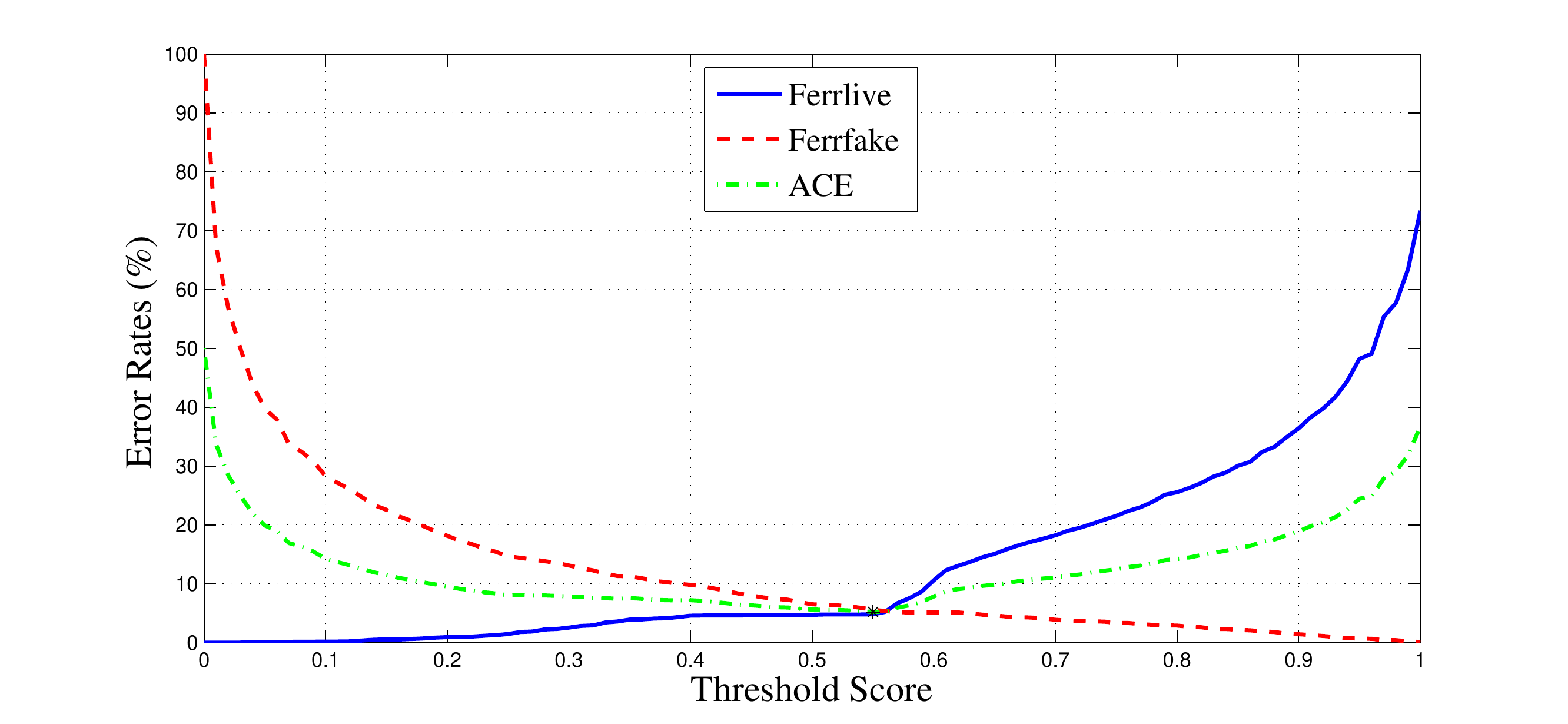}		
	}}
	\resizebox{0.49\textwidth}{!}{
		\subfigure[]
		{ 		
			\includegraphics[width=18cm, height=14cm]{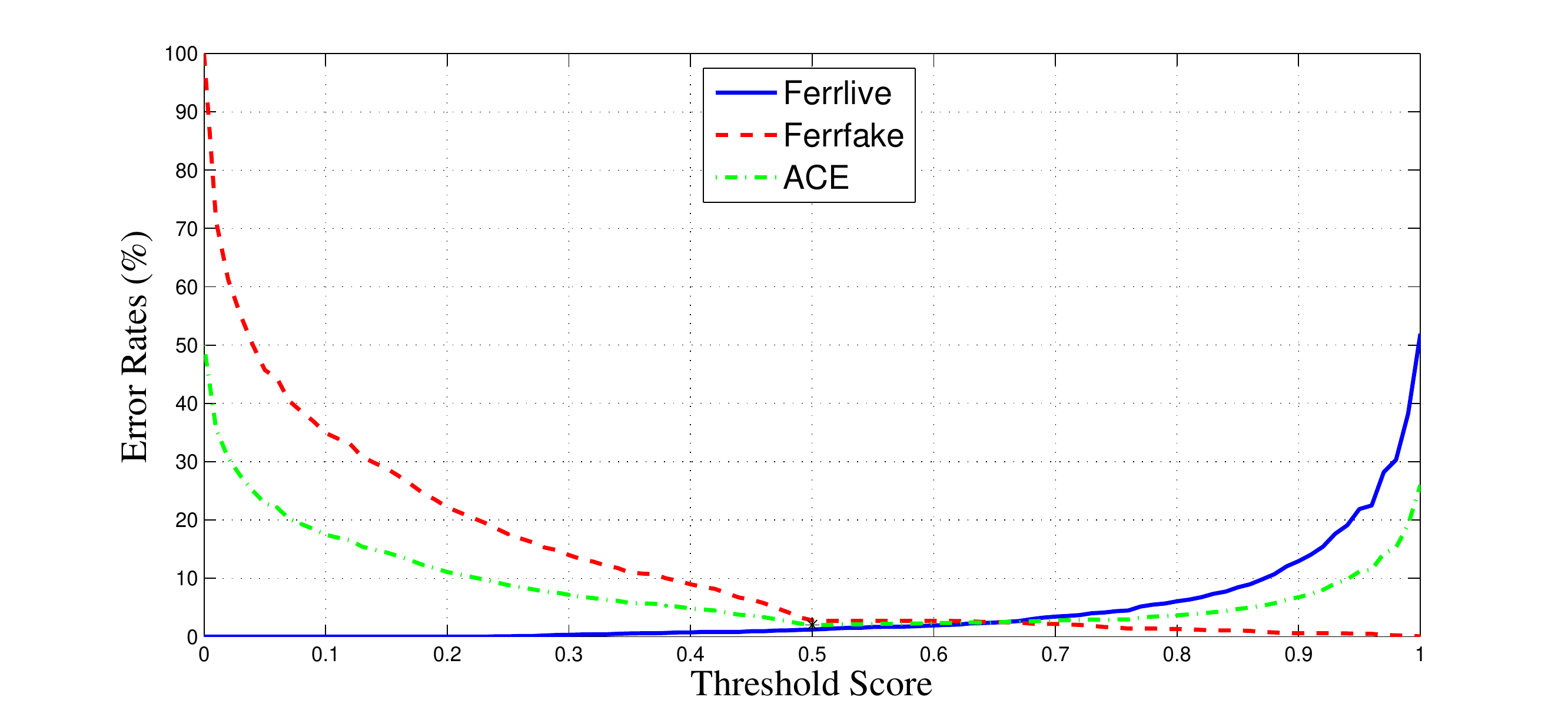}		
	}}
	\caption{FerrLive and FerrFake rates at different threshold of probability score for (a)  Biometrika (b) CrossMatch (c) Identix sensors. The star mark represents the least average classification error.  }
	\vspace{-4mm}
	\label{ErrorCurve}
\end{figure}

\begin{table}[b]
	\centering
	\caption{FerrFake, FerrLive, and ACE rates (in \%) at best, EER, and LivDet threshold }
	\label{ThresholdErrorRates}
	\resizebox{\textwidth}{!}{
		\begin{tabular}{@{}cccccccccc@{}}
			\toprule
			\multirow{2}{*}{} & \multicolumn{3}{c}{Biometrika}    & \multicolumn{3}{c}{CrossMatch}      & \multicolumn{3}{c}{Identix}         \\ \cmidrule(l){2-4} \cmidrule(l){5-7} \cmidrule(l){8-10}
			& Best & EER  & LiveDet  & Best & EER  & LiveDet & Best & EER & LiveDet \\ \cmidrule(l){1-10}
			Threshold         & 0.59        & 0.56 &     0.50           & 0.55        & 0.56 &       0.50         & 0.50        & 0.66 &      0.50          \\ 
			ACE               & 7.9        & 8.3 &    8.5            & 5.2        & 5.3 &      5.6          & 2.0         & 2.5 &      2.0          \\
			Ferrfake          & 9.1        & 8.2 &     10.2           & 4.8        & 5.2 &    6.5            & 1.2         & 2.5 &        2.8        \\
			Ferrlive          & 6.7        & 8.3 &       6.8         & 5.6        & 5.3 &      4.7          & 2.8         & 2.4 &      1.2          \\
			\bottomrule
	\end{tabular}}
	\vspace{-4mm}
\end{table}
It is observed that, the test fingerprints having probability score above 0.6 for live and below 0.4 for fake is correct while majority of the incorrectly classified fingerprints lies in between 0.4 and 0.6. In our experiments, we have selected 0.5 as the cutoff score (as used in LivDet competition) to make a decision whether it is live or fake fingerprint instead of selecting an optimal threshold where Ferrfake and Ferrlive rates can be minimized. Therefore, to decide the optimal threshold, Ferrfake and Ferrlive rates are calculated at the different threshold of probability score in the range of 0 to 1 with a step size of 0.01. Fig. \ref{ErrorCurve} shows the Ferrfake, Ferrlive, and ACE rates at the different threshold of probability scores for Biometrika, Crossmatch, and Identix sensors. The intersection point of the Ferrfake and Ferrlive curve shows equal error rate (EER) point while the best ACE is shown using * on ACE curve (green). Comparison of the error rates at best, EER, and the LivDet threshold of probability score is shown in Table \ref{ThresholdErrorRates}. For Biometrika sensor, ACE rates are $7.9\%$ at best threshold (0.59), $8.3\%$ at EER point (0.56), and $8.5 \%$ at LivDet threshold (0.50). The ACE rate of $7.9$ shows that changing the cutoff threshold appropriately can minimize the error rates of the liveness detection system. The ACE rate for Crossmatch sensors is $5.2 \%$ at best threshold ($0.55$), $5.3 \%$ at EER threshold ($0.56$), and $5.6\%$ at LiveDet threshold. It can be verified from the Fig. \ref{ErrorCurve} (b) that the ACE at EER point is almost the same as best ACE. For Identix sensor, the best ACE remains $2.0 \%$ at LivDet threshold (0.50) while the ACE of $2.5 \%$ is achieved at EER threshold (0.66) as seen from the Fig. \ref{ErrorCurve} (c).

\begin{table}[b]
	\centering
	\caption{Comparative results in terms of Ferrlive, Ferrfake, and ACE (in \%) for LivDet 2009.}
	\label{comparison}
	\resizebox{\textwidth}{!}{
		\begin{tabular}{lllllllllll}
			\toprule
			\multirow{3}{*}{}                                          & \multicolumn{10}{c}{Comparative Results: LiveDet 2009}                                                   \\   \midrule
			& \multicolumn{3}{c}{Biometrika} & \multicolumn{3}{c}{CrossMatch} & \multicolumn{3}{c}{Identix} &             \\ \cmidrule(r){2-4} \cmidrule(l){5-7} \cmidrule(l){8-10}
			& Ferrlive   & Ferrfake   & ACE  & Ferrlive   & Ferrfake   & ACE  & Ferrlive  & Ferrfake  & ACE & \begin{tabular}[c]{@{}l@{}}Average \\ ACE\end{tabular} \\ \midrule 
			Abhyankar et. al. \cite{Abhyankar2006}&    24.2        &    39.2        &    31.7  &     39.7       &     23.3       &   31.5   &  48.4         &    46.0       &  47.2   &     36.8        \\  
			Nikam et. al \cite{Nikam2010}&   14.3         &   42.3         &   28.3   &      19.0      &    18.4        &   18.7   &    23.7       &      37.0     &   30.3  &   25.8          \\
			Moon et. al. \cite{Moon2005}&    20.8        &   25.0         &  23.0    &     27.4       &    19.6        &  23.5    &    74.7       &     1.6      &  38.2   &         28.2    \\
			Marsco et. al. \cite{MARASCO2012}&     12.2       &   13.0         &  12.6    &    17.4        &    12.9        &  15.2    &  8.3         &     11.0      & 9.7    &     12.5        \\
			Best LivDet09 \cite{LivDet09}&    15.6        &    20.7        &  18.2    &      7.4      &    11.4        &  9.4    &     2.7      &      2.8     &   2.8  &    10.1         \\	
			Galbally et. al \cite{GALBALLY2012} &    \textbf{3.1}        &    71.8        &  37.4    &    8.8        &    20.8        &   14.8   &     4.8      &    5.0       &   4.9  &     19.0        \\	
			IQA-based \cite{Galbally2014} &     14.0       &   11.6         &  12.8    &     8.6       &   12.8         &   10.7   &   \textbf{1.1}        &      \textbf{1.4}     &  \textbf{1.2}   &   8.2          \\							
			
			Xia et. al \cite{Xia2017}&    -        &    -        &  11.3    &     -       &      -      &  5.4    &     -      &    -       &   2.0  &    6.2         \\

			Proposed method &    6.8        &     \textbf{10.2}       &   \textbf{8.5}  &     \textbf{4.7}       &     \textbf{6.5}       &   \textbf{5.6}   &     1.2     &  2.8         &  2.0   &      \large\textbf{5.3 }      \\ \bottomrule        
	\end{tabular}}
\end{table}
	\vspace{-5mm}
\subsection{Comparision with existing approaches}
	\vspace{-2mm}
Several research works have been carried out in the field of fingerprint liveness detection to identify the fake fingerprints. Here, we compare the achieved results with the results mentioned in \cite{Galbally2009,Xia2017} for different liveness detection algorithms. From the comparison results reported in Table \ref{comparison}, it is evident that the proposed approach outperforms the existing methods in two of the datasets (Biometrika and CrossMatch). However, the performance slightly deteriorates than IQA \cite{Galbally2014} for Identix sensor but it is still comparable. Generally, the overall performance of the fingerprint liveness detection system should be well across all the sensors and fake fingerprint fabrication materials. The overall ACE (average of Biometrika, Crossmatch, and Identix sensor) of the proposed method is least (5.3\%), which signifies that our approach is capable of adapting to different sensors and different fake fingerprint fabrication materials. 
\begin{figure}[b]
	\centering	
	\resizebox{0.49\textwidth}{!}{
		\subfigure[]
		{ 		
			\includegraphics[width=18cm, height=14cm]{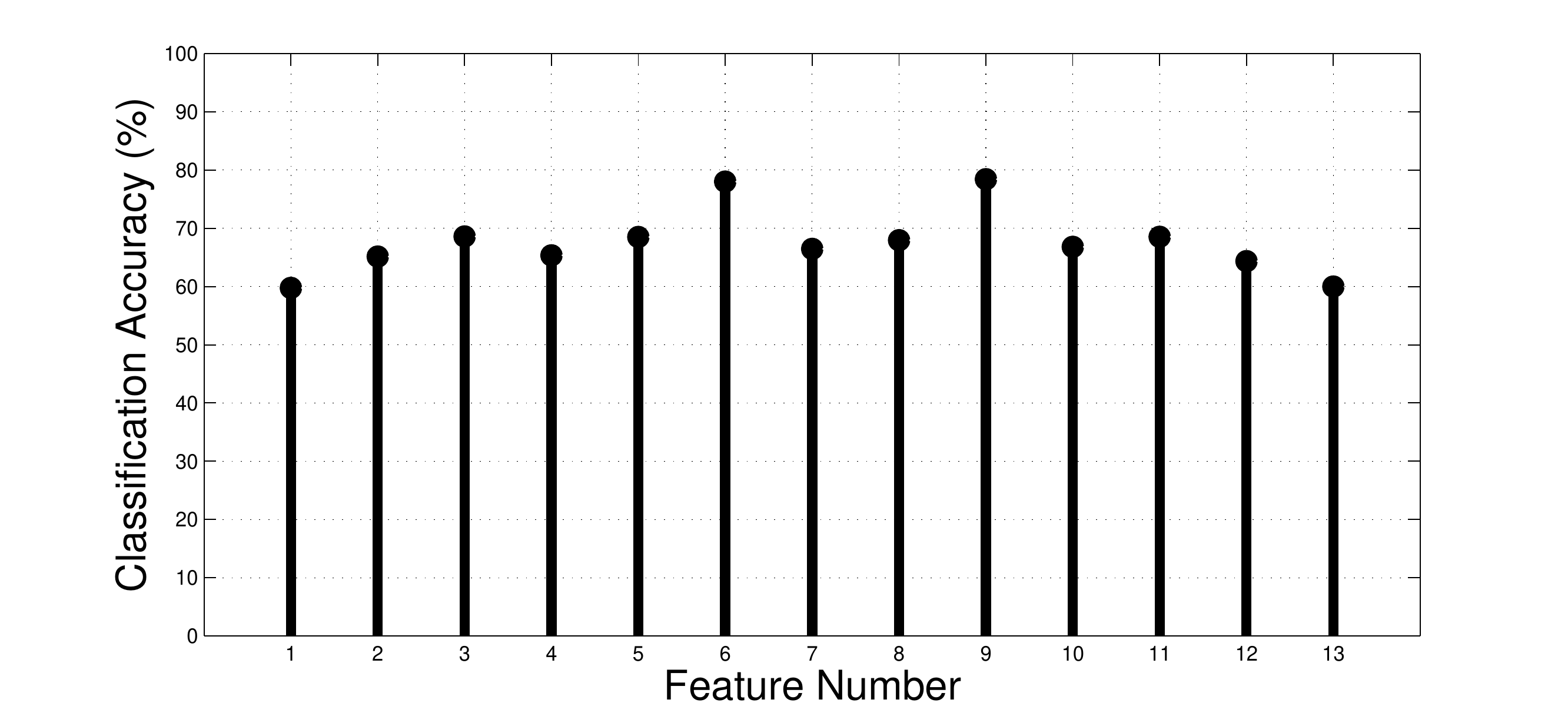}		
	}}
	\resizebox{0.49\textwidth}{!}{
		\subfigure[]
		{ 		
			\includegraphics[width=18cm, height=14cm]{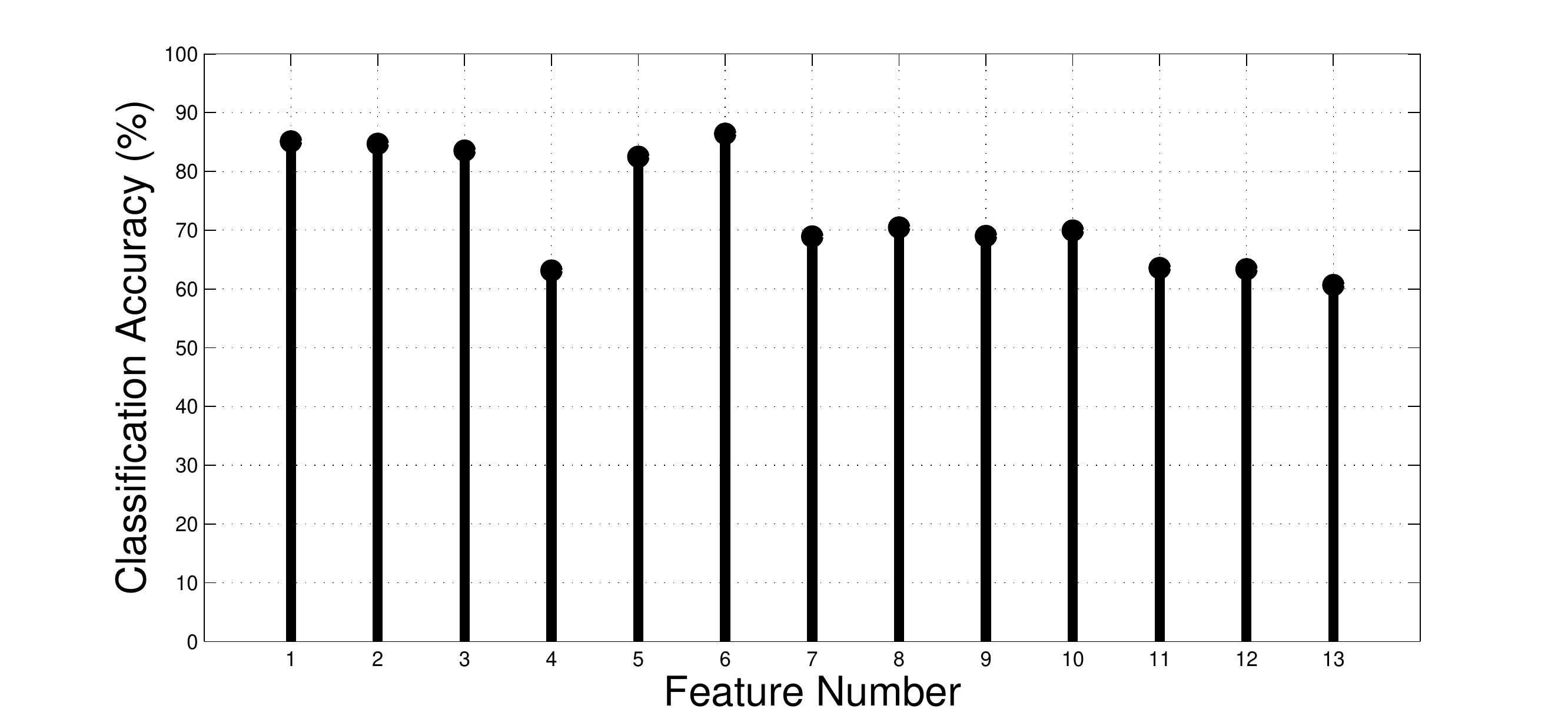}		
	}}
	\resizebox{0.49\textwidth}{!}{
		\subfigure[]
		{ 		
			\includegraphics[width=18cm, height=14cm]{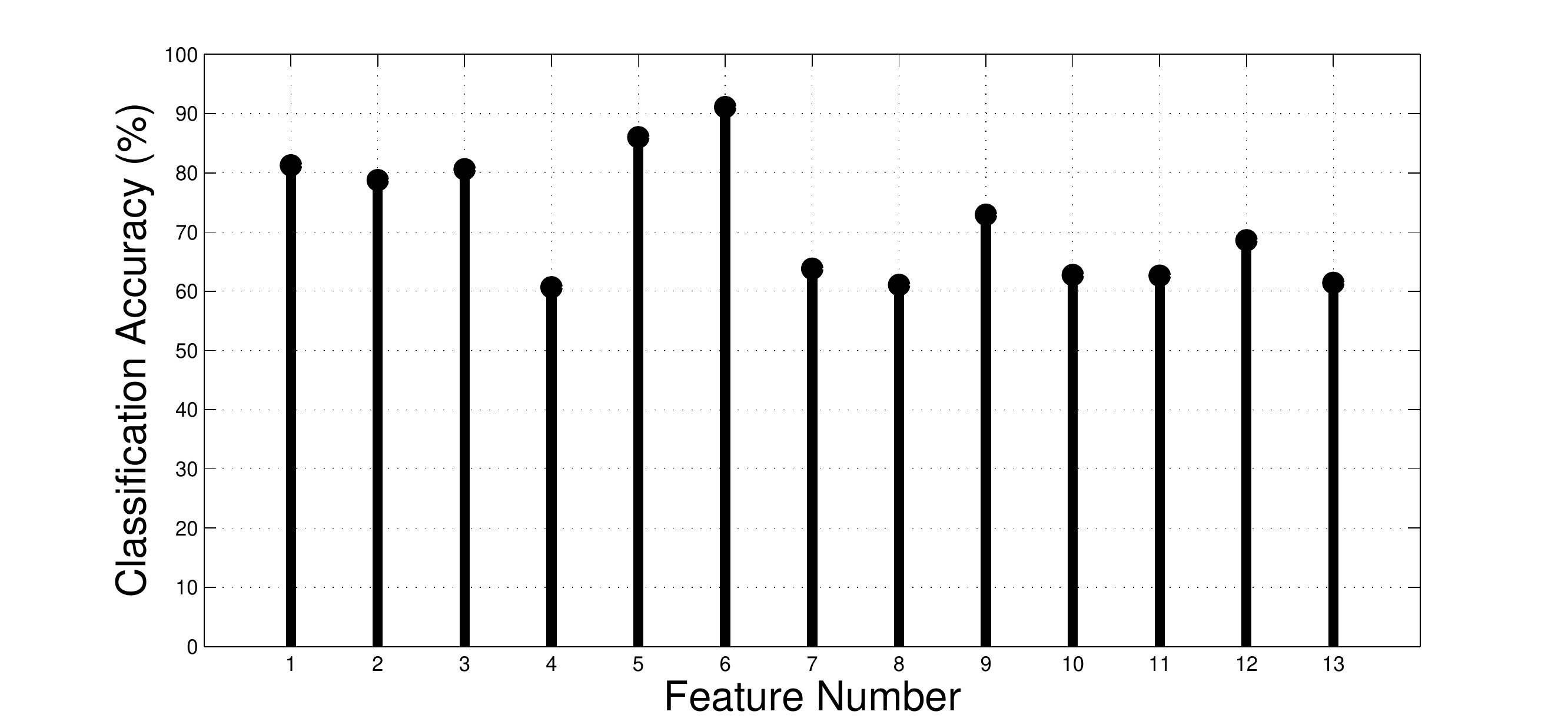}		
	}}
	\caption{Classification accuracy using the individual features as feature vector for (a)  Biometrika (b) CrossMatch (c) Identix sensors }
	
	\label{FeatureAnalysis}
\end{figure}

\begin{table}[t]
	\vspace{-2mm}
	\centering
	\caption{Strong features for each sensor including overall best features}
	\label{StrongFeature}
	\begin{tabular}{cc}
		\hline
		Sensors          & Strong Features                                                               \\ \hline
		Biometrika       & $V_{ab}^\mu$, $FDA^\mu$                                                     \\
		Crossmatch       & $RWS^\mu$, $RWS^\sigma$, $VWS^\mu$, $R_{ab}^\mu$, $V_{ab}^\mu$            \\
		Identix          & $RWS^\mu$, $RWS^\sigma$, $VWS^\mu$,  $R_{ab}^\mu$, $V_{ab}^\mu$           \\
		\textbf{Overall} & $RWS^\mu$, $RWS^\sigma$, $VWS^\mu$, $R_{ab}^\mu$, $V_{ab}^\mu$, $FDA^\mu$ \\ \hline
	\end{tabular}
	\vspace{-4mm}
\end{table}
	\vspace{-5mm}
\subsection{Feature individuality analysis} 
	\vspace{-2mm}
In this section, a preliminary study is carried out to determine the discriminative power of the individual features used in the proposed liveness detection system. For this purpose, we consider the features mentioned in section \ref{featurevector} as an individual feature to assess their classification performance. As per the features reported in Table \ref{featureSelection}, the classification performance using individual feature is illustrated in Fig. \ref{FeatureAnalysis}. The features with accuracies higher than 75\% for each sensor are considered as strong features. These features can be used to assess fingerprint liveness in other datasets. Table \ref{StrongFeature} summarizes strong features for each sensor of LivDet 2009 datasets which constitutes overall best feature set for fingerprint liveness detection.

\vspace{-5mm}

\subsection{Performance of strong features for unknown spoof materials}
	\vspace{-2mm}
In LivDet 2015 datasets, test set contain fingerprints fabricated using unknown spoof materials which are not part of training set. Therefore, effectiveness of the overall strong features identified over LivDet 2009 in Table \ref{StrongFeature} is evaluated on LivDet 2015 datasets. ACE of the proposed method along with the ACE of the winner of LiveDet 2015 is shown in Table \ref{LivDet2015}. The proposed method achieves an overall ACE of $4.22 \%$ with respect to the $4.49 \%$ of the winner of LivDet 2015, which confirms the robustness of the proposed approach for liveness detection over unknown spoof materials and different sensors. 

\begin{table}[h]
	\centering
	\caption{Performance comparison of error rates (in \%) between the proposed approach (bottom) and LivDet 2015 winner (top)}
	\label{LivDet2015}
	\begin{tabular}{lllll}
		\hline
		& LivDet 2015     & Ferrlive & Ferrfake & ACE  \\ \hline
		& GreenBit        &   3.50       &   5.30       &  4.40   \\
		& Biometrika      &  8.50        &   3.73       &  5.64   \\\textbf{LivDet 2015 winner}
		& Digital Persona &   8.10       &     5.07     &  6.28   \\
		& Crossmatch     &   0.93         &    2.90      &  1.90   \\
		& Average         &     4.78     &     4.27     &  \textbf{4.49}   \\ \hline
		
        & GreenBit        &   4.03       &    4.64      &  4.33   \\
		& Biometrika      &    7.10      &    2.46      &  4.78   \\\textbf{Proposed method}
		& Digital Persona &     6.83     &     4.85     &  5.84   \\
		& Crossmatch      &    1.34      &     2.53     &   1.93  \\
		& Average         &  4.82       &   3.62       & \textbf{4.22}   \\ \hline
	\end{tabular}
\end{table}
\vspace{-5mm}

 \section{Conclusions and future work}
 \vspace{-2mm}
In this paper, we have proposed a novel method for fingerprint liveness detection. Owing to the dispersion and variations in the ridge-valley structure of the fake fingerprint images, we have proposed novel quality based features (i.e. RWS, VWS, $R_{ab}$, $V_{ab}$, RVC). These features explore the minute details of ridge-valley structure discriminating real and fake fingerprints. The proposed technique combines multiple features extracted from a single image to detect liveness. First, we perform experimental evaluation  on LivDet 2009 datasets and obtained an overall ACE of $5.3\%$. In-depth comparison shows that the proposed method outperforms the current state-of-the-art liveness detection approaches for LivDet 2009 Datasets. Performance of the proposed features is also determined individually to find out the best performing features. The effectiveness of the overall best features is experimentally tested on LivDet 2015. The ACE rate of $4.22 \%$ in comparison to the $4.49 \%$ of the LivDet 2015 winner is a significant performance improvement.  

Our experimental results demonstrate that the proposed quality feature based method is able to handle various spoofing materials and sensors consistently well over different datasets. Therefore, generality and robustness of the proposed approach is adequate for liveness detection of different spoof materials. Additionally, as proposed method requires a single image, it is more user-friendly, faster and computationally efficient. In future work, proposed liveness detection system can be integrated with quality assessment module of fingerprint recognition system. Effectiveness of other quality related features can also be evaluated for fingerprint liveness detection.

\vspace{-6mm}

\section*{Compliance with ethical standards} 
\vspace{-2mm}
Funding: This research is supported by the Science $\&$ Engineering Research Board (SERB) grant number ECR/2017/000027.\\
Conflict of Interest: Second author of this paper has received research grants from Science $\&$ Engineering Research Board (SERB) and declares no conflict of interest.
\vspace{-4mm}
\begin{acknowledgements}
The authors are thankful to SERB (ECR/2017/000027), Department of science  Technology, Govt. of India for providing financial support. Also, We would like to acknowledge Indian Institute of Technology Indore for providing the laboratory support and research facilities to carry out this research work.
\end{acknowledgements}
\vspace{-4mm}
\bibliographystyle{spmpsci}      



\end{document}